\documentclass[letterpaper]{article} 
\usepackage{aaai2026}  
\usepackage{times}  
\usepackage{helvet}  
\usepackage{courier}  
\usepackage[hyphens]{url}  
\usepackage{graphicx} 
\usepackage{natbib}  
\usepackage{caption} 
\usepackage{algorithm}
\usepackage{algorithmic}

\usepackage{amsmath,amssymb,amsfonts}
\usepackage{booktabs}
\usepackage{multirow}
\usepackage{array}
\usepackage{adjustbox}

\usepackage{newfloat}
\usepackage{listings}
\DeclareCaptionStyle{ruled}{labelfont=normalfont,labelsep=colon,strut=off} 
\floatstyle{ruled}
\newfloat{listing}{tb}{lst}{}
\floatname{listing}{Listing}
\title{Early Warning of Intraoperative Adverse Events via Transformer-Driven Multi-Label Learning}
\author {
    Xueyao Wang\textsuperscript{\rm 1,\rm 2},
    Xiuding Cai\textsuperscript{\rm 1,\rm 2},
    Honglin Shang\textsuperscript{\rm 1,\rm 2},
    Yaoyao Zhu\textsuperscript{\rm 3}, 
    Yu Yao\textsuperscript{\rm 1,\rm 2}\thanks{Corresponding author}
}
\affiliations {
    \textsuperscript{\rm 1}Chengdu Institute of Computer Application, Chinese Academy of Sciences, Chengdu 610213, China\\
    \textsuperscript{\rm 2}University of Chinese Academy of Sciences, Beijing 100049, China\\
    \textsuperscript{\rm 3}China Zhenhua Research Institute Co., Ltd., Guiyang 550014, China\\
    
    \{wangxueyao221, caixiuding20, shanghonglin24, zhuyaoyao19\}@mails.ucas.ac.cn, casitmed2022@163.com
}

\usepackage{bibentry}

\begin{document}

\maketitle

\begin{abstract}

Early warning of intraoperative adverse events plays a vital role in reducing surgical risk and improving patient safety. While deep learning has shown promise in predicting the single adverse event, several key challenges remain: overlooking adverse event dependencies, underutilizing heterogeneous clinical data, and suffering from the class imbalance inherent in medical datasets. To address these issues, we construct the first Multi-label Adverse Events dataset (MuAE) for intraoperative adverse events prediction, covering six critical events. Next, we propose a novel Transformer-based multi-label learning framework (IAENet) that combines an improved Time-Aware Feature-wise Linear Modulation (TAFiLM) module for static covariates and dynamic variables robust fusion and complex temporal dependencies modeling. Furthermore, we introduce a Label-Constrained Reweighting Loss (LCRLoss) with co-occurrence regularization to effectively mitigate intra-event imbalance and enforce structured consistency among frequently co-occurring events. Extensive experiments demonstrate that IAENet consistently outperforms strong baselines on 5, 10, and 15-minute early warning tasks, achieving improvements of +5.05\%, +2.82\%, and +7.57\% on average F1 score. These results highlight the potential of IAENet for supporting intelligent intraoperative decision-making in clinical practice.
\end{abstract}


\section{Introduction}

Surgery is a cornerstone of modern healthcare, with over 300 million procedures performed annually worldwide~\cite{nepogodiev2019global}. However, it carries a disproportionately high risk of harm: 46–65\% of all medical adverse events are surgery-related, and 3–22\% of surgical patients experience complications~\cite{meara2015global}. Intraoperative adverse events are common, when prolonged, and are associated with severe complications such as pulmonary dysfunction, acute kidney injury, and cardiovascular events. These outcomes contribute to higher mortality and worse long-term prognosis~\cite{varghese2024artificial}. Importantly, most are preventable with timely intervention, underscoring the urgent need for accurate risk early warning systems.

\begin{figure}
    \centering
    \includegraphics[width=1\linewidth]{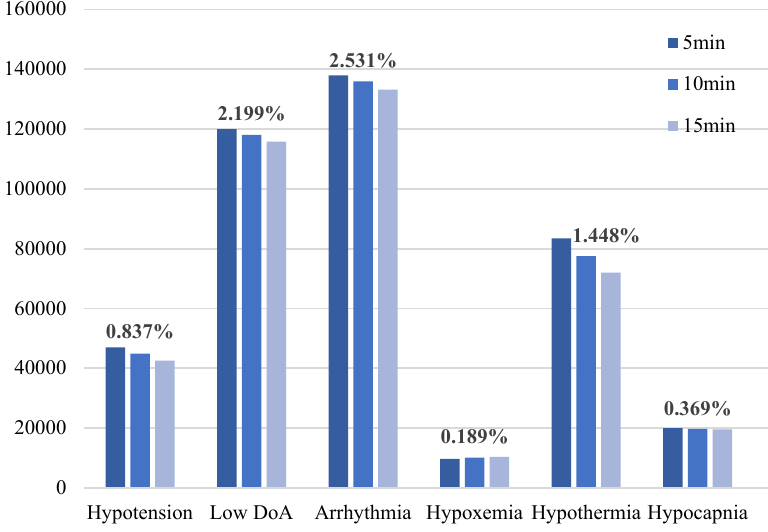}
    \caption{Positive sample distribution by adverse event for 5, 10, and 15 minutes prediction windows in MuAE. Percentages indicate the event occurrence rate among total samples.}
    \label{fig:pos-neg label}
\end{figure}

Advanced predictive models leveraging Artificial Intelligence (AI) could improve perioperative care by identifying high-risk patients, enabling early interventions, and ultimately reducing preventable harm~\cite{cai2024outcomes, cai2025advances}. Currently, the mainstream approach to perioperative risk modeling focuses on single adverse event prediction using machine learning and deep learning, including tasks like intraoperative Hypotension (IOH)~\cite{chen2021forecasting, lu2023composite, moon2024frequency}, Hypoxemia~\cite{lundberg2018explainable, park2023machine}, inadequate and excessive Depth of Anesthesia (DoA) prediction~\cite{lee2018prediction, lee2019data}. ~\citeauthor{lee2021deep} used a 1D Convolutional Neural Network for regression and classification tasks to predict intraoperative hypotension events using invasive and non-invasive arterial pressure, thereby reducing postoperative organ dysfunction risks. ~\citeauthor{chen2021forecasting} merely extended this paradigm to multiple events by training separate models for each outcome, albeit with shared self-supervised features. Prevailing single-event approaches overlook the intrinsic dependencies between adverse events. While modeling multiple events jointly can improve individual predictions by leveraging their mutual relationships, existing methods fail to explicitly model the dependencies within the time-series modality. Moreover, the paradigm shift still faces the key challenge: the intra-event imbalance from just 0.189\% to 2.531\% of the total samples (see~\textbf{Fig.~\ref{fig:pos-neg label}}), severely hinders model generalization. Despite its clinical relevance, multi-adverse-event prediction remains largely unexplored in perioperative risk assessment.

Besides, recent studies have increasingly focused on integrating static covariates and dynamic variables~\cite{schnider2021relationship, bahador2021multimodal, lu2023composite, moon2024frequency} through various fusion strategies. For instance, ~\citeauthor{moon2024frequency} integrated dynamic features extracted from both time and frequency domains with static variables via simple concatenation for downstream classification. Meanwhile,~\citeauthor{lu2023composite} fused dynamic variables with static counterparts into four distinct modalities, employing attention mechanisms for long-range interaction in hypotension prediction. However, such direct fusion strategies often introduce feature redundancy and noise, thereby increasing the complexity of representation learning and hindering model performance.

To bridge existing gaps, we construct the first intraoperative \textbf{Mu}ltiple \textbf{A}dverse \textbf{E}vent (MuAE) dataset based on the public VitalDB dataset to address the research gap in early warning of multi-adverse events. Then, we propose an \textbf{I}ntraoperative \textbf{A}dverse \textbf{E}vents \textbf{Net}work (IAENet) for multi-label time-series classification. It deploys a \textbf{T}ime-\textbf{A}ware \textbf{F}eature-wise \textbf{L}inear \textbf{M}odulation (TAFiLM) module on the transformer encoder to dynamically modulate static covariates and dynamic variables for better feature fusion to reduce redundant noise. The transformer encoder is used to extract the temporal feature for the classification task, where the inverted embedding of variables as input tokens inherently captures multivariate correlations. Additionally, we also design a \textbf{L}abel-\textbf{C}onstrained \textbf{R}eweighting Loss (LCRLoss), which dynamically reweights prediction outputs based on batch-wise label frequency and incorporates a co-occurrence regularization term to model structured label dependencies, mitigating intra-event imbalance and improving model robustness and generalization. To summarize, our main contributions are the following: 
\begin{itemize}
    \item We construct the first multi-label dataset for early warning of intraoperative adverse events, the MuAE dataset, comprising six adverse event types derived from the VitalDB dataset.
\end{itemize}
\begin{itemize}
    \item We propose a transformer-based model for intraoperative multi-adverse events classification, named IAENet. It employs a Time-Aware Feature-wise Linear Modulation (TAFiLM) module for modulation fusion of static covariates and dynamic vital signs to enhance feature quality and reduce redundant noise.
\end{itemize}
\begin{itemize}
    \item We design a Label Constraint Reweighting Loss function (LCRLoss) to effectively mitigate the intra-event imbalance problem and capture structured label dependencies.
\end{itemize}
\begin{itemize}
    \item Extensive experiments and ablation studies demonstrate the effectiveness of the proposed framework.
\end{itemize}

\section{Related Works}
\subsubsection{Intraoperative Single-event Early Warning.}
AI-driven early warning for intraoperative adverse events enables earlier clinical intervention, enhancing patient safety and improving surgical outcomes~\cite{cai2024outcomes}. Existing research~\cite{lee2018prediction, lundberg2018explainable, hwang2023intraoperative, park2023machine, lu2023composite, moon2024frequency} predominantly focuses on single-event prediction through time-series forecasting or classification. For instance,~\citeauthor{hwang2023intraoperative} predicted the 
event solely from arterial blood pressure waveforms, employing discrete wavelet transforms and CNN-based feature extraction.~\citeauthor{lundberg2018explainable} used ensemble models with $\mathrm{SpO_2}$-derived features for hypoxemia prediction, incorporating risk factor analysis for interpretability. Such isolated event prediction fails to capture perioperative complexity comprehensively and neglects clinically significant interdependencies between co-occurring adverse events. To address the current gap in multi-adverse-event prediction research, we propose the first Transformer-based multi-label learning model that explicitly captures inter-event dependencies to enhance predictive performance.

\subsubsection{Class Imbalance in Multi-label Learning.}
Medical datasets typically suffer from class imbalance, where sustained normal states outweigh brief adverse events. In multi-label settings, the imbalance is more complex due to interdependent label distributions. For example, bradycardia often co-occurs with hypotension~\cite{cheung2015predictors}, highlighting intra-event interaction in multi-label intraoperative prediction. To mitigate this, prior work explores cost-sensitive learning~\cite{lin2017focal, cao2019learning} and label-aware resampling~\cite{cui2019class}. However, synthetic signals like SMOTE~\cite{chawla2002smote} may distort physiological patterns, while static loss weights like Asymmetric loss~\cite{ridnik2021asymmetric} fail to adapt to temporal label dynamics or capture co-occurrence. We address these issues with a label-constrained reweighting loss that models inter-label dependencies and adapts to distributional shifts during training.

\subsubsection{Medical Time Series Modeling.} 
Medical time-series data from heterogeneous monitoring devices often exhibit misalignment, noise susceptibility, and multiple variables. Deep learning models have been employed to capture the temporal dependencies and inter-variable correlations in time series analysis~\cite{wang2024deep}. With the inclusion of different covariates and vital signs data, multivariate data present challenges in effective integration. Recent models like iTransformer~\cite{liu2024itransformer} leverage attention mechanisms to model temporal and variable-level dependencies jointly. However, many existing methods concatenate static variables (expanded over time steps) with dynamic inputs directly, which introduces redundancy and hampers model efficiency. So we improve the FiLM module~\cite{perez2018FiLM} based on the transformer to perform dynamic modulation of vital signs series and static variables for better early feature fusion.

\section{Dataset}
Due to the lack of publicly available datasets on intraoperative multiple adverse events, we develop the MuAE dataset by processing the VitalDB dataset~\cite{lee2022vitaldb} through rigorous data cleaning and preprocessing. VitalDB contains anesthesia records from 6,388 non-cardiac surgeries between August 2016 and June 2017 at Seoul National University Hospital, encompassing comprehensive perioperative physiological signals and anesthetic parameters.

\subsection{Dataset Cleaning}
Case selection was performed first. Inclusion and exclusion criteria were used to ensure data quality through selection criteria guided by clinical experts. Cases were selected for surgical durations of more than 2 hours to ensure adequate monitoring data under general anesthesia. Paediatric cases were excluded if they were aged less than 18 years and weighed less than 35 kg. Patients with an ASA classification higher than grade 6 were also excluded.

Then, vital signs and static features were selected from the intraoperative monitoring data. 15 key physiological dynamic variables are filtered from raw monitoring characteristics, such as drug infusion parameters (PPF20\_VOL, RFTN20\_VOL, PPF20\_CE, and RFTN20\_CE), neuromonitoring (BIS), blood pressure (ART\_DBP, ART\_MBP, ART\_SBP), temperature (BT), heart rate (HR), oxygen saturation (PLETH\_SPO2) and ECG (ECG\_II) can be filtered. Patient's age, weight, height, gender, and ASA classification were selected as 5 static covariates.

Finally, the 15 vital signs variables were resampled at 2-second intervals. Negative or blank values were set to zero and treated as missing data. Missing values were filled using forward and backward interpolation to minimize information loss. For anesthetic drug infusion parameters, we converted the drug unit volume to a rate for indirect use. The final selection of 873 patients was made after data quality control. We spliced 15 dynamic variables $\mathbf{x}_{d}$ and 5 static covariates $\mathbf{x}_{s}$ as network inputs $\mathbf{x}$:

\begin{equation}\mathbf{x}=
\begin{pmatrix}
 \\
\underbrace{\mathbf{x}_{d_0},\cdots,\mathbf{x}_{d_{14}}}_{\text{Dynamic variables}}, \underbrace{\mathbf{x}_{s_0},\cdots,\mathbf{x}_{s_4}}_{\text{Static covariates}}
\end{pmatrix} \in\mathbb{R}^{W \times (D+S)}, 
\end{equation}
where $W$ represents the input window size and $D$, $S$ represent the number of dynamic variables and static covariates.

\begin{figure}
    \centering
    \includegraphics[width=1\linewidth]{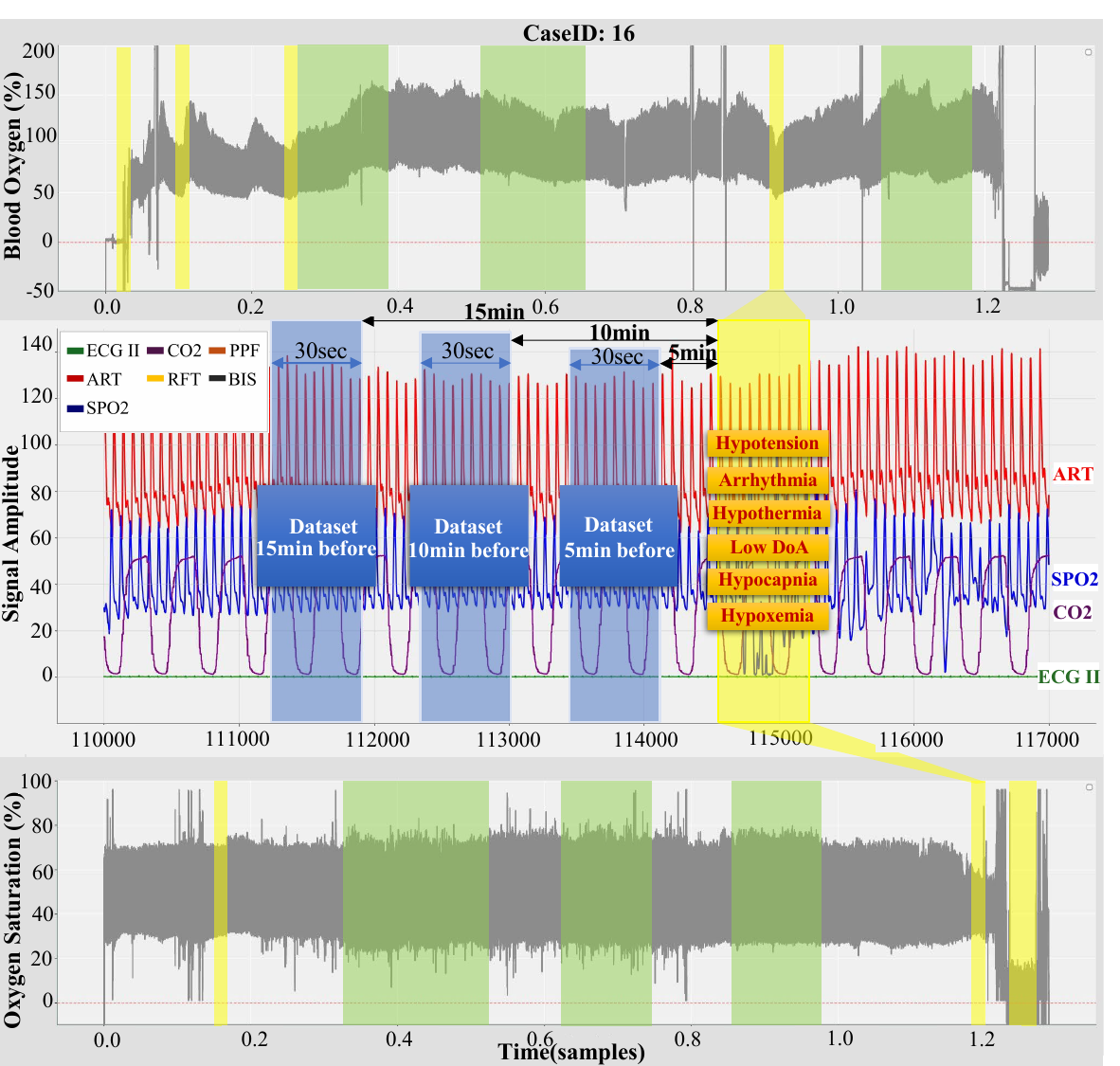}
    \caption{The pipeline of Multi-adverse Events Early Warning}
    \label{fig: pipeline}
\end{figure}

\begin{figure*}[ht]
    \centering
    \includegraphics[width=1\textwidth]{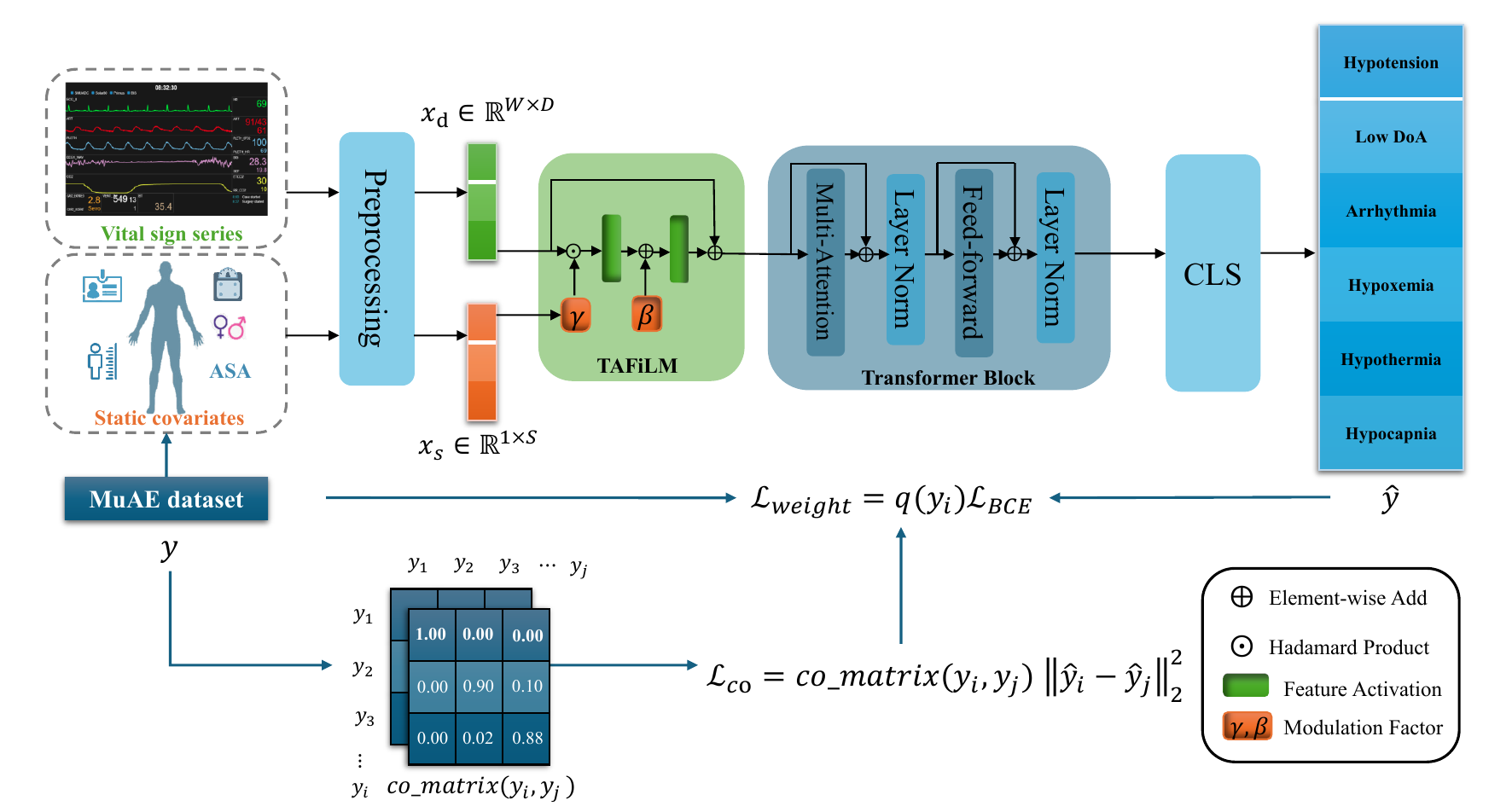}
    \caption{An overview of the IAENet framework for time-series in multi-label classification. Given an sample about vital sign series and static covariate in MuAE dataset $\mathbf{x} = \{\mathbf{x}_{d_0},..., \mathbf{x}_{d_{14}}, \mathbf{x}_{s_0},..., \mathbf{x}_{s_4} \}$, our goal is to predict whether the values in the following $\triangle$ time steps will be normal or abnormal $\mathbf{y} = \{\mathbf{y}_{1}, \mathbf{y}_{2}, ..., \mathbf{y}_{6} \}$. Firstly, the preprocessed dynamic variables and static covariates are respectively fused through the TAFiLM module for feature early fusion. Then, a Transformer encoder is employed to capture temporal correlations among multivariate variables. Finally, the model is trained using the proposed LCRLoss, which combines a batch-wise label frequency-weighted BCE loss with a co-occurrence constraint term based on label dependencies.}
    \label{fig: the IAENet model structure}
\end{figure*}

\subsection{Event Label Definition}
Six major intraoperative adverse events were defined based on clinical guidelines and prior studies~\cite{chen2021forecasting, lu2023composite, moon2024frequency}: hypotension, low depth of anesthesia, arrhythmia, hypoxemia, hypothermia, and hypocapnia. Event occurrence $\mathbf{y}_{t+\Delta:t+\Delta+T}^i \in \{0, 1\}^{6}$ was determined when monitored parameters exceeded predefined thresholds: 

\begin{itemize}
    \item Hypotension: MAP $<$ 65 mmHg for at least 1 minute.
\end{itemize}
\begin{equation}\max(\mathbf{x}_\mathrm{MAP}[t+\Delta:t+\Delta+30])<65.\end{equation}
\begin{itemize}
    \item Low depth of anesthesia: BIS $<$ 40 for at least 1 minute.
\end{itemize}
\begin{equation}\max(\mathbf{x}_\mathrm{BIS}[t+\Delta:t+\Delta+30])<40.\end{equation}
\begin{itemize}
    \item Arrhythmia: HR $<$ 60 bpm or HR $>$ 100 bpm for at least 1 minute.
\end{itemize}
\begin{equation}
\begin{split}&\left(\max(\mathbf{x}_\mathrm{HR}[t+\Delta:t+\Delta+30]) < 60 \right) \\
&\quad \lor \left(\min(\mathbf{x}_\mathrm{HR}[t+\Delta:t+\Delta+30]) > 100 \right).
\end{split}\end{equation}
\begin{itemize}
    \item Hypoxemia: $\mathrm{SpO_{2}}$ $<$ 90\% for at least 1 minute.
\end{itemize}
\begin{equation}\max(\mathbf{x}_\mathrm{SpO_{2}}[t+\Delta:t+\Delta+30])<90.\end{equation}
\begin{itemize}
    \item Hypothermia: BT $<$ 35°C for at least 1 minute.
\end{itemize}
\begin{equation}\max(\mathbf{x}_\mathrm{BT}[t+\Delta:t+\Delta+30])<35.\end{equation}
\begin{itemize}
    \item Hypocapnia: $\mathrm{EtCO_{2}}  <$ 30 mmHg for at least 1 minute.
\end{itemize}
\begin{equation}\max(\mathbf{x}_\mathrm{EtCO_{2}}[t+\Delta:t+\Delta+30])<30.\end{equation}

Given a continuous $\triangle$-minute sequence of physiological signals preceding time point $t$, we train a multi-label classification model $f_\theta$ to predict whether any of six representative intraoperative adverse events will occur during the subsequent $T$-minute interval. We define the predicted outcome $\mathbf{\hat{y}}_i = \mathbf{\hat{y}}_{t+\Delta:t+\Delta+T}^i$ as the output of the network function:

\begin{equation}
\mathbf{\hat{y}}_{t+\Delta:t+\Delta+T}^i=f_\theta(\mathbf{x}[t-W:t]),
\end{equation}
where $\triangle$ represents the advance prediction time, $T$ represents the predicted event window length, $i$ represents the adverse events, and $t$ represents the current timespoint.

\section{Methods}
\subsection{Data Preprocessing}
The multivariate physiological data were normalized to eliminate scale differences caused by inconsistent feature units. The data were then segmented into overlapping time windows of 30 seconds in length, using a 2-second sliding step. Approximately 4.88 million, 5.66 million, and 5.55 million samples were generated under the 5, 10, and 15 minutes ahead prediction settings, respectively.

The~\textbf{Fig.~\ref{fig: pipeline}} shows the pipeline of multi-adverse events early warning. Physiological data segments are labeled as normal (green) or abnormal (yellow). Input data are extracted from 30-second intervals preceding the occurrence of abnormal events (blue) within 5/10/15-minute windows in advance. The output data is presented using binary labels. The top and bottom panels demonstrate label generation for hypotension (blood pressure) and hypoxemia (oxygen saturation) events, highlighting distinct patterns between hemodynamic and respiratory instability. 

\subsection{Time Series Classification Model}
The overview of the proposed IAENet is shown in~\textbf{Fig.~\ref{fig: the IAENet model structure}}. It integrates a Transformer encoder for multivariate temporal modeling and a TAFiLM module to fuse static covariates with dynamic vital signs through conditional modulation. The input data $\mathbf{x}$ is derived from the MuAE dataset. After preprocessing, the features are first passed through the TAFiLM module for early fusion. The fused representations are then fed into the Transformer encoder to capture multivariate temporal dependencies. During training, the model adopts a label-constrained reweighting loss as the core optimization strategy.

\subsubsection{TAFiLM.}
Inspired by the Feature-wise Linear Modulation (FiLM) module~\cite{perez2018FiLM}, we propose the Time-Aware FiLM (TAFiLM) module for time series data. The FiLM module enables conditional feature modulation through learnable affine transformations, allowing dynamic adjustment of feature representations based on specific contextual conditions. While FiLM has demonstrated strong performance in multimodal tasks, its application to time series data remains underexplored. 

To achieve effective feature fusion, the static conditional features $\mathbf{x}_{s} \in \mathbb{R}^{B \times S}$ are processed by a conditional network to generate time-varying scaling and shifting factors $\gamma_s, \beta_s \in \mathbb{R}^{B \times W \times D}$. The parameters dynamically modulate the dynamic features through affine transformation within the TAFiLM module, formulated as: 

\begin{equation}
\mathrm{TAFiLM}(\mathbf{\mathbf{x}_{d}})=((\gamma_s+1)\odot \mathbf{x}_{d} +\beta_s) \in \mathbb{R}^{B \times W \times D},
\end{equation}
where $\odot$ represents the hadamard product, $\mathbf{x}_{d} \in \mathbb{R}^{B \times W \times D}$ represents the dynamic feature, and the conditional network chooses the MLP layer.

\subsubsection{Transformer Block.}
Following in time-series tasks~\cite{liu2024itransformer}, we reverse the input sequence before token embedding, addressing the timestamp misalignment issues in standard Transformer architectures. After embedding, the tokens are applied via a multi-head self-attention mechanism in the encoder-only transformer, capturing the dynamic interactions of the variables. To ensure variable independence, each variable is individually normalized by the LayerNorm. Finally, they are processed individually using a feed-forward network to create a sequential representation for the downstream classification task. 

\subsubsection{LCRLoss Definition.}
To mitigate label imbalance, we propose a batch-wise reweighting strategy that scales loss weights inversely to label frequencies. For batch $b$ with $N$ samples, the weighted BCE loss is:

\begin{equation}\mathcal{L}_{weight}=\frac{1}{C}\sum_{c=1}^{C}\left(q_{pos}^{b}\sum_{i\in\mathcal{P}_{c}}\mathcal{L}_{BCE}+q_{neg}^{b}\sum_{i\in\mathcal{N}_{c}}\mathcal{L}_{BCE}\right),
\end{equation}
where $\mathcal{P}c$ and $\mathcal{N}c$ denote positive and negative samples for class $C$, and weights $q_{pos}^b$, $q_{neg}^b$ are updated per batch. Standard BCE loss is used for validation and testing.

Among the evaluated frequency-based reweighting strategies, the square root inverse scheme is adopted due to its overall stability, as demonstrated in the ablation study.

\begin{equation}q(\mathbf{y}_i)=
    \begin{cases}
        \frac{1}{sqrt(\mathcal{P}c / N)} & \mathrm{if}\ \mathbf{y}_i = 1, \\
        \frac{1}{sqrt(\mathcal{N}c / N)} & \mathrm{otherwise}. 
    \end{cases}
\end{equation} 

To overcome the independence assumption in per-class reweighting, we introduce a co-occurrence loss $\mathcal{L}_{co}$ that encourages consistency among labels that frequently co-occur.

Firstly, we compute a label co-occurrence matrix $\mathbf{M}$ from the entire training set, capturing the correlations among adverse events. Assume the training set contains $N$ samples, which are divided into $B$ batches. The b-th batch contains $n_b$ samples. The computation is formally expressed as:

\begin{equation}
\mathbf{M}=\sum_{b=1}^B (\mathbf{y}^{(b)})^\top \mathbf{y}^{(b)}, \mathbf{M}\in \mathbb{R}^{C \times C}.
\end{equation}

Secondly, the aggregated co-occurrence matrix $\textit{co\_matrix}$ is further normalized by its maximum value to ensure numerical stability and mitigate scale discrepancies:

\begin{equation}
\textit{co\_matrix}=\frac{\mathbf{M}}{\max_{i,j}\mathbf{M}_{ij}+\epsilon},
\end{equation}
where $\epsilon$ is a small constant added to prevent division by zero. The co-occurrence matrix $\mathbf{M}$ is symmetric, i.e., $\mathbf{M}_{ij} = \mathbf{M}_{ji}$.

To enforce consistency among frequently co-occurring labels, we then compute the squared L2 distance between all label prediction pairs ($\mathbf{\hat{y}}_{i}$, $\mathbf{\hat{y}}_{j}$), weighted by the corresponding normalized co-occurrence value $\textit{co\_matrix}$($\mathbf{y}_i$, $\mathbf{y}_j$). The co-occurrence regularization loss, $\mathcal{L}_{co}$ is defined as:

\begin{equation}
\mathcal{L}_{co}=\sum_{i,j}^C\textit{co\_matrix}(\mathbf{y}_i,\mathbf{y}_j)\|\mathbf{\hat{y}}_{i}-\hat{\mathbf{y}}_{j}\|_2^2,
\end{equation}
where $C$ is the total number of labels, $\mathbf{\hat{y}}_{i}$ and $\mathbf{\hat{y}}_{j}$ are the prediction of the model (e.g., logits or probabilities) for label $\mathbf{y}_i$ and $\mathbf{y}_j$, respectively.

Finally, the overall loss $\mathcal{L}_{LCR}$ is composed of two components, balanced by a weighting coefficient $\lambda$:
\begin{equation}
\mathcal{L}_{LCR}=\mathcal{L}_{weight}+\lambda \mathcal{L}_{co}.
\end{equation}

\begin{table*}[ht]

  \centering
  \setlength{\extrarowheight}{2pt} 
  \scriptsize 
  \setlength{\tabcolsep}{1mm}
  \begin{tabular}{cc cc cc cc cc cc cc cc cc cc cc cc}
    \toprule
    \multicolumn{2}{c}{\multirow{1}{*}{\textbf{Models}}} & 
    \multicolumn{2}{c}{\multirow{1}{*}{IAENet}} &
    \multicolumn{2}{c}{\multirow{1}{*}{iTransformer}} &
    \multicolumn{2}{c}{\multirow{1}{*}{Crossformer}} &
    \multicolumn{2}{c}{\multirow{1}{*}{PatchTST}} &
    \multicolumn{2}{c}{\multirow{1}{*}{FEDformer}} &
    \multicolumn{2}{c}{\multirow{1}{*}{Autoformer}} &
    \multicolumn{2}{c}{\multirow{1}{*}{Informer}} &
    \multicolumn{2}{c}{\multirow{1}{*}{DLinear}} &
    \multicolumn{2}{c}{\multirow{1}{*}{Non-stationary}} &
    \multicolumn{2}{c}{\multirow{1}{*}{TFT}} &
    \multicolumn{2}{c}{\multirow{1}{*}{SegRNN}} \\ 
    
    \toprule
    
    & \multicolumn{1}{c}{$\triangle$} & F1 & AUC & F1 & AUC & F1 & AUC & F1 & AUC & F1 & AUC & F1 & AUC & F1 & AUC & F1 & AUC & F1 & AUC & F1 & AUC & F1 & AUC \\
    
    \cmidrule{2-2} \cmidrule(lr){3-4} \cmidrule(lr){5-6} \cmidrule(lr){7-8} \cmidrule(lr){9-10} \cmidrule(lr){11-12} \cmidrule(lr){13-14} \cmidrule(lr){15-16} \cmidrule(lr){17-18} \cmidrule(lr){19-20} \cmidrule(lr){21-22} \cmidrule(lr){23-24}

    \multirow{3}{*}{\rotatebox{90}{Hypotension}} 
    
    & 5 & \textbf{51.45} & \textbf{74.83} & 45.64 & 65.76 & \underline{48.32} & 67.29 & 24.30 & 57.58 & 
    45.90 & 66.69 & 3.02 & 50.69 & 47.65 & 67.96 & 47.65 & \underline{67.69} & 48.37 & 68.66 & 36.71 & 62.59 & 29.23 & 59.19 \\
    
    & 10 & \textbf{36.21} & \textbf{65.85} & 16.91 & 54.79 & 19.14 & 55.50 & 10.58 & 52.80 & 
    11.59 & 53.11 & 1.07 & 50.10 & \underline{25.16} & \underline{58.35} & 17.81 & 55.20 & 23.10 & 57.40 & 19.09 & 55.54 & 14.47 & 54.55 \\
    
    & 15 & \textbf{28.54} & \textbf{62.08} & 9.79 & 52.57 & 15.24 & 54.28 & 5.45 & 51.29 & 
    10.13 & 52.67 & 0.03 & 49.95 & 11.17 & 53.02 & 8.01 & 52.03 & 16.10 & 54.84 & \underline{16.81} & \underline{55.31} & 10.17 & 52.68 \\
    
    \midrule

    \multirow{3}{*}{\rotatebox{90}{Low DoA}}
    
    & 5 & \textbf{61.45} & \textbf{76.69} & 54.03 & 69.49 & 37.70 & 66.77 & 2.28 & 50.43 & 
    51.42 & 68.08 & 6.11 & 50.31 & 49.53 & 67.19 & 49.53 & 67.19 & 47.92 & 66.19 & 6.18 & 51.06 & \underline{54.65} & \underline{70.35} \\
    
    & 10 & \textbf{56.54} & \textbf{73.42} & \underline{49.80} & \underline{67.35} & 43.18 & 63.80 & 0.95 & 50.17 & 
    43.85 & 64.14 & 11.62 & 50.57 & 37.89 & 61.17 & 43.77 & 64.09 & 31.35 & 58.56 & 7.08 & 51.21 & 43.59 & 64.01 \\
    
    & 15 & \textbf{50.21} & \textbf{68.16} & 28.95 & 57.81 & 39.50 & 62.00 & 0.96 & 50.16 & 
    30.91 & 58.48 & 14.71 & 50.73 & 27.19 & 56.53 & 28.87 & 57.85 & 30.18 & 57.74 & 8.90 & 50.72 & \underline{32.34} & \underline{59.07} \\
    
    \midrule

    \multirow{3}{*}{\rotatebox{90}{Arrhythmia}}
    
    & 5 & \textbf{65.04} & \textbf{77.40} & 62.76 & 75.19 & \underline{63.68} & \underline{76.05} & 5.13 & 51.02 & 
    56.29 & 70.63 & 3.79 & 50.00 & 52.35 & 68.47 & 52.35 & 68.47 & 59.93 & 73.43 & 26.23 & 56.17 & 60.82 & 73.76 \\
    
    & 10 & \textbf{59.12} & \textbf{73.77} & \underline{57.91} & \underline{72.28} & 56.22 & 70.72 & 5.76 & 51.06 & 
    55.15 & 70.08 & 21.97 & 51.76 & 51.06 & 67.86 & 0.67 & 50.12 & 47.84 & 66.01 & 15.77 & 52.60 & 57.28 & 72.00 \\
    
    & 15 & \textbf{54.79} & \textbf{71.04} & 48.36 & 66.28 & \underline{53.47} & \underline{69.86} & 3.01 & 50.52 & 
    48.99 & 66.58 & 2.08 & 50.19 & 41.08 & 62.51 & 0.57 & 50.11 & 41.94 & 62.92 & 15.71 & 51.79 & 51.78 & 68.37 \\

    \midrule

    \multirow{3}{*}{\rotatebox{90}{Hypoxemia}} 
    
    & 5 & \textbf{57.79} & \textbf{88.25} & \underline{56.34} & 75.68 & 49.65 & 71.57 & 55.18 & 77.22 & 
    51.81 & 73.41 & 1.98 & 50.44 & 55.88 & \underline{77.95} & 49.83 & 69.86 & 51.90 & 74.62 & 56.17 & 76.80 & 46.51 & 69.69 \\
    
    & 10 & \textbf{51.56} & \textbf{84.77} & \underline{45.58} & \underline{68.70} & 43.61 & 66.96 & 36.95 & 63.43 & 
    34.39 & 61.67 & 3.88 & 50.99 & 39.25 & 66.17 & 36.56 & 63.02 & 42.59 & 69.55 & 35.98 & 62.92 & 33.19 & 61.28 \\
    
    & 15 & \textbf{43.65} & \textbf{81.37} & 17.07 & 55.08 & 30.49 & 61.14 & 15.66 & 54.55 & 
    24.60 & 58.15 & 0.00 & 50.00 & 24.01 & 59.03 & 17.19 & 55.08 & 19.66 & 56.45 & \underline{28.79} & \underline{63.36} & 21.02 & 56.75 \\
    
    \midrule

    \multirow{3}{*}{\rotatebox{90}{Hypothermia}} 
    
   & 5 & \textbf{84.64} & \textbf{92.77} & 72.40 & 82.69 & \underline{71.22} & \underline{84.67} & 34.97 & 61.02 & 
    75.19 & 85.12 & 3.06 & 50.59 & 70.62 & 81.79 & 70.62 & 81.80 & 67.19 & 81.84 & 42.41 & 64.63 & 77.19 & 86.17 \\
    
    & 10 & \textbf{75.59} & \textbf{86.38} & \underline{73.64} & \underline{86.19} & 60.08 & 73.05 & 31.20 & 59.31 & 
    59.67 & 74.01 & 6.64 & 51.00 & 54.17 & 72.79 & 47.64 & 66.55 & 51.95 & 70.07 & 35.97 & 61.37 & 65.28 & 79.05 \\
    
    & 15 & \textbf{59.93} & \textbf{77.00} & 37.96 & 62.37 & 45.13 & 65.48 & 22.95 & 56.27 & 
    42.61 & 64.52 & 1.50 & 50.25 & 41.47 & 65.72 & 33.03 & 60.13 & 36.28 & 62.43 & 30.87 & 59.37 & \underline{51.36} & \underline{71.18} \\

    \midrule

    \multirow{3}{*}{\rotatebox{90}{Hypocapnia}} 
    
    & 5 & \textbf{42.29} & \textbf{75.31} & 36.49 & 62.21 & \underline{37.95} & \underline{63.04} & 19.26 & 55.83 & 
    28.80 & 59.03 & 1.96 & 50.39 & 37.98 & 65.84 & 37.98 & 65.84 & 35.87 & 64.31 & 2.34 & 50.59 & 30.51 & 60.96 \\
    
    & 10 & \textbf{28.17} & \textbf{62.14} & 7.80 & 52.04 & 5.22 & 51.32 & 2.33 & 50.57 & 
    1.43 & 50.34 & 0.56 & 50.09 & 6.76 & 51.74 & 10.25 & 52.74 & \underline{13.52} & \underline{54.13} & 7.47 & 51.95 & 4.42 & 51.11 \\
    
    & 15 & \textbf{25.89} & \textbf{63.04} & 1.29 & 50.29 & 2.02 & 50.47 & 0.36 & 50.08 & 
    6.19 & 51.58 & 0.10 & 50.01 & 8.13 & 52.16 & 2.48 & 50.58 & 11.31 & 53.21 & \underline{11.27} & \underline{53.17} & 3.21 & 50.79 \\

    \toprule
    \toprule

    \multirow{3}{*}{\rotatebox{90}{Mean}} 
    & 5 & \textbf{65.36} & \textbf{80.86} & 60.06 & \underline{71.84} & \underline{60.31} & 71.56 & 16.24 & 58.85 & 57.51 & 70.49 & 4.24 & 50.41 & 54.76 & 71.54 &41.03 & 64.57 & 56.21 & 71.51 & 26.30 & 60.31 & 59.07 & 70.02 \\
    
    & 10 & \textbf{58.51} & \textbf{74.39} & \underline{55.69} & \underline{66.89} & 48.94 & 63.58 & 12.40 & 54.56 & 48.23 & 62.22 & 13.66 & 50.75 & 44.60 & 63.02 & 30.41 & 58.62 & 41.01 & 62.62 & 19.04 & 55.93 & 51.00 & 63.58 \\
    
    & 15 & \textbf{50.52} & \textbf{70.45} & 35.74 & 57.40 & \underline{42.95} & \underline{60.45} & 7.67 & 52.14 & 37.67 & 58.66 & 6.62 & 50.19 & 33.64 & 58.16 & 19.00 & 54.30 & 33.64 & 57.93 & 17.72 & 55.62 & 41.82 & 59.81 \\
    
    \bottomrule
  \end{tabular}%
  \caption{Classification Accuracy results across all tasks and methods. $\triangle$ denotes the forecasting horizon time steps. All metrics are in \%. The best results are in \textbf{bold} font and the second best are \underline{underlined}.}
  \label{tab:full_forecasting_results}
\end{table*}

\section{Experiments}
\subsubsection{Baseline Methods.}
To assess the classification capabilities of the compared methods, we evaluate the IAENet with MLP-based model DLinear~\cite{zeng2023transformers}; RNN-based model SegRNN~\cite{lin2023segrnn}; Transformer-based models iTransformer~\cite{liu2024itransformer}, PatchTST~\cite{nie2022time}, FEDformer~\cite{zhou2022fedformer}, Autoformer~\cite{wu2021autoformer}, Informer~\cite{zhou2021informer}, Temporal Fusion Transformer (TFT)~\cite{yeche2023temporal}, Crossformer~\cite{zhang2023crossformer}, and Non-stationary Transformer~\cite{liu2022non}.

\subsubsection{Implementation Details.}
IAENet was developed primarily using the Time-Series-Library framework~\cite{wu2023timesnet}. The samples were divided according to patients into training (70\%), validation (10\%), and test (20\%) sets for model validation. We used an RAdam optimizer with an initial learning rate of $1e^{-3}$. The optimal value of $\lambda$ determined by grid search was set to 0.02, where the search range is [0.001, 0.01, 0.02, 0.05, 0.1, 0.2]. The batch size was set to 64 for 10 epochs with an early stopping patience of 3 epochs. The IAENet utilized the proposed LCRLoss, while all other models used the standard BCE Loss.

\subsubsection{Evaluation Metrics.}
To evaluate the performance of the IAENet in the multi-label classification task, we employed six commonly used metrics: Micro F1 (F1), Macro AUC (AUC), Micro Precision (PRE), Micro Recall (REC), Micro Accuracy (ACC), and Hamming Loss (HM). Among them, F1 and AUC serve as the primary evaluation indicators, as they jointly reflect both prediction accuracy and completeness, and maintain robustness under class imbalance.

\subsection{Experimental Results}
\subsubsection{Performance Comparison.}
To evaluate the effectiveness of IAENet, we designed experiments inspired by previous studies~\cite{lee2021deep, yang2024dynamic, moon2024frequency}, conducting multi-adverse event prediction tasks at 5, 10, and 15 minutes in advance. Among six adverse event classifications, the IAENet consistently outperforms all baseline models in F1 and AUC, as shown in~\textbf{Table~\ref{tab:full_forecasting_results}}. Compared to the competitive Crossformer and iTransformer models, IAENet achieves average F1 score improvements of 5.05\%, 2.82\%, and 7.57\% across the three classification tasks, respectively. Performance declines as input time intervals increase, indicating that the windows near event onset contain more discriminative features for accurate classification. Additionally, certain adverse events show consistently poor performance across all baselines, likely due to severe label imbalance.

\subsubsection{Comparison with Different Loss.}
To evaluate the effectiveness of the LCRLoss, we compared it against several representative loss functions designed for imbalanced data, including weighted binary cross-entropy (WBCE), binary cross-entropy (BCE), Focal Loss (FL), Polynomial Loss (PolyLoss), Asymmetric Loss (ASL), and SoftMarginLoss (SMLoss). The comparative results are presented in~\textbf{Table~\ref{tab3}}. The results show that LCRLoss achieves the best performance among all compared loss functions. Compared to the most competitive alternative ASL, it improves average F1 score by 0.49\% and AUC by 1.02\%, demonstrating its effectiveness in handling intra-label imbalance.

\begin{table}
    
    \centering
    \small
    \setlength{\tabcolsep}{1.9mm}
    \begin{tabular}{lcccccc}
        \toprule
        Loss & ACC & PRE & REC & \textbf{F1} & \textbf{AUC} & HM  \\
        \midrule
        
        LCRLoss & 91.72 & 61.77 & 69.39 & \textbf{65.36} & \textbf{80.88} & 0.0828 \\
        WBCE & 92.17 & 68.22 & 56.99 & 62.10 & 71.12 & 0.0783   \\
        BCE & 92.69 & 71.70 & 57.83 & 64.03 & 74.21 & 0.0731 \\
        FL & \textbf{92.73} & \textbf{73.76} & 54.91 & 62.96 & 73.75 & \textbf{0.0727}\\ 
        ASL & 91.22 & 58.98 & \textbf{72.08} & \underline{64.87} & \underline{79.86} & 0.0878  \\ 
        PolyLoss & 92.66 & 70.90 & 58.91 & 64.35 & 75.07 & 0.0734  \\ 
        SMLoss & 92.59 & 71.96 & 55.93 & 62.94 & 73.65 & 0.0741  \\
        
        \bottomrule
    
    \end{tabular}
    \caption{Performance of IAENet under different losses.}
    \label{tab3}
\end{table}

\subsubsection{Ablation Analysis.}
To evaluate the effectiveness of TAFiLM and LCRLoss, we integrated them into several baseline models. In the baselines, static covariates and dynamic variables are directly concatenated as input and trained using the BCE loss. As shown in~\textbf{Table~\ref{tab5}}, adding only FiLM, TAFiLM, and LCRLoss improves the average F1 and AUC across baselines, indicating that the components act as effective plug-and-play modules. Notably, TAFiLM outperforms FiLM, indicating its superior ability to modulate dynamic features through static inputs while mitigating noise from direct concatenation. TAFiLM slightly reduces precision but significantly improves recall, which is important in medical diagnosis to minimize underdiagnosis.

\begin{table}
\centering
    \small
    \setlength{\tabcolsep}{1.6mm}
    \begin{tabular}{lccccccc}
        \toprule
        Module & ACC & PRE & REC & \textbf{F1} & \textbf{AUC} & HM  \\
        \midrule
        
        Informer & 91.28 & 65.74 & 46.92 & 54.76 & \textbf{71.54} & 0.0872\\
        + LCRLoss & 90.24 & 55.87& \textbf{62.89} & \textbf{59.17} & \textbf{79.36} & 0.0976 \\
        + FiLM & \textbf{92.05} & \textbf{70.10} & 51.14 & \underline{59.13} & 70.62 & \textbf{0.0795} \\
        + TAFiLM & 91.95 & 67.89 & 53.96 & \textbf{60.13} & \underline{71.34} & 0.0805  \\ 
        \midrule
        Crossformer & 92.39 & \textbf{73.03} & 51.36 & \underline{60.31} & \textbf{71.56} & 0.0761  \\
        + LCRLoss & 90.80 & 57.88 & \textbf{66.80} & \textbf{62.02} & \textbf{81.17} & 0.0920\\
        + FiLM & 92.21 & 71.11 & 51.36 & 59.99 & 71.46 & 0.0779\\ 
        + TAFiLM & \textbf{92.49} & 72.45 & 53.63 & \textbf{61.63} & \underline{71.55} & \textbf{0.0751} \\
        \midrule
        SegRNN & 91.84 & 67.86 & 52.29 & 59.07 & 70.02 & 0.0816 \\
        + LCRLoss & 90.93 & 58.11 & \textbf{69.51} & \textbf{63.30} & \textbf{80.50} & 0.0907\\
        + FiLM & 92.30 & \textbf{72.40} & 50.99 & \underline{59.83} & \underline{70.74} & 0.0770\\
        + TAFiLM & \textbf{92.34} & 70.84 & 54.26 & \textbf{61.45} & \textbf{71.40} & \textbf{0.0766} \\
        \midrule
        iTransformer & 92.19 & 70.76 & 52.16 & 60.06 & 71.84 & 0.0781  \\
        + LCRLoss & 91.02 & 59.48 & \textbf{63.31} & \textbf{61.34} & \textbf{77.99} & 0.0898\\
        + FiLM & 92.46 & \textbf{72.90} & 52.55 & \underline{61.07} & \underline{71.86} & 0.0754   \\ 
        + TAFiLM & \textbf{92.69} & 71.70 & 57.83 & \textbf{64.03} & \textbf{74.21} &\textbf{ 0.0731}  \\
        
        \bottomrule
    
    \end{tabular}
    \caption{Ablation studies on the components of IAENet.}
    \label{tab5}
\end{table}

We further conducted a comprehensive ablation study to assess the impact of different reweighting strategies, batch size, and the coefficient $\lambda$ on model performance, as measured by F1 and AUC (see~\textbf{Table~\ref{tab:ablation_f1_auc}}). We first evaluated several reweighting schemes in LCRLoss, including inverse, log inverse, square root inverse, and cube root inverse on global frequency and batch-wise local frequency. The batch-wise frequency-based $sqrt\_inverse$ strategy achieves the highest F1, indicating a strong balance between precision and recall. The inverse and log inverse strategies achieve high AUC but suffer from low F1, suggesting overfitting or instability. Overall, batch-wise label frequency-based reweighting consistently outperforms global static reweighting, indicating the importance of incorporating local data distributions under imbalanced label distributions. Based on the trade-off, the $sqrt\_inverse$ is selected as the default. 

Considering that the representativeness based on batch statistics may be sensitive to the batch size, we then computed a co-occurrence matrix from training data to obtain stable estimates of adverse event correlations. Results demonstrate that global co-occurrence statistics computed on the full training set more accurately reflect event correlations than small batches of estimates. Furthermore, we examined the effect of the hyperparameter $\lambda$, which balances the contributions of the BCE loss and the co-occurrence loss.

\begin{table}[t]
\centering
\small
\setlength{\tabcolsep}{1.4mm}

\begin{tabular}{llcc}
\toprule
Factor & Setting & \textbf{F1} & \textbf{AUC} \\
\midrule
\multirow{5}{*}{Weighting (G / L)} 
    & none           &  63.97       &  74.30        \\
    & inverse        & 40.55 / 60.82                & 78.80 / \textbf{85.35} \\
    & log\_inverse   & 24.40 / 51.02                & 60.23 / \underline{83.94} \\
    & sqrt\_inverse  & 59.51 / \textbf{65.36}       & \textbf{83.68} / 80.88 \\
    & cubic\_inverse & \textbf{63.91} / \underline{64.26}    & 82.33 / 79.44         \\
\midrule
\multirow{4}{*}{Batch Size} 
    & 64   & \underline{65.21} & 81.35 \\
    & 128  & 64.64 & \underline{81.38}        \\
    & 256  & 64.78             & \textbf{81.76}    \\
    & All  & \textbf{65.36}    & 80.88             \\
\midrule
\multirow{5}{*}{\textit{$\lambda$}} 
    & 0.001  & 62.28   & 72.58    \\
    & 0.01  & 64.89    & 75.06    \\
    & 0.02  & \textbf{65.36}    & \textbf{80.88}      \\
    & 0.05  & 63.33    & 74.41   \\
    & 0.1  & 62.43     & 73.2    \\ 
    
\bottomrule
\end{tabular}
\caption{Ablation studies on reweighting strategies, batch size, and $\lambda$ of LCRLoss.}
\label{tab:ablation_f1_auc}
\end{table}

\subsubsection{Gradient Analysis.}
\begin{figure}[ht]
    \centering
    \includegraphics[width=1.\linewidth]{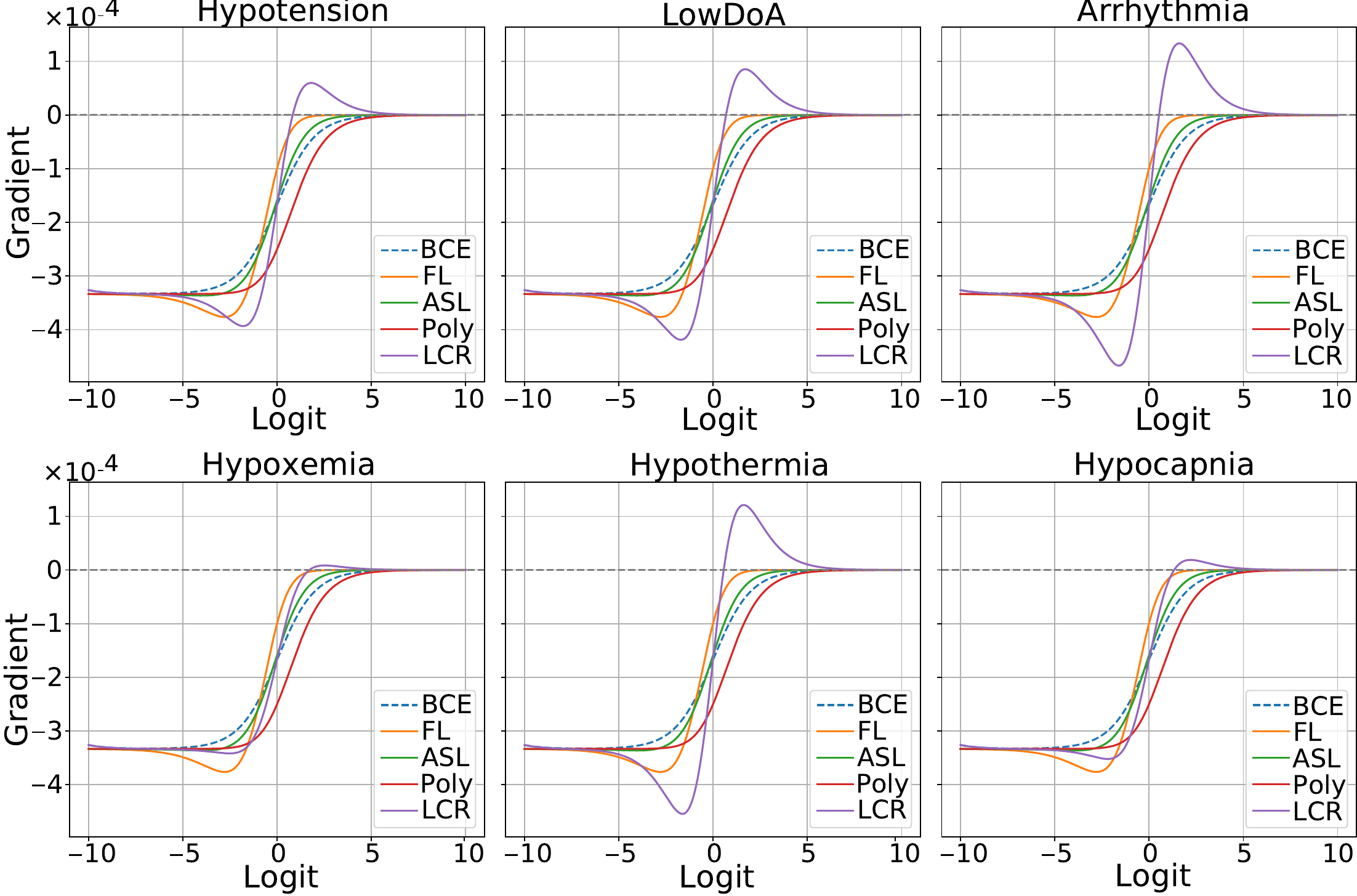}
    \caption{Derivatives of the loss functions. The X-axis denotes the logit of positive labels, and the Y-axis is the corresponding gradients.}
    \label{fig: Gradients}
\end{figure}

\begin{figure}
    \centering
    \includegraphics[width=0.95\linewidth]{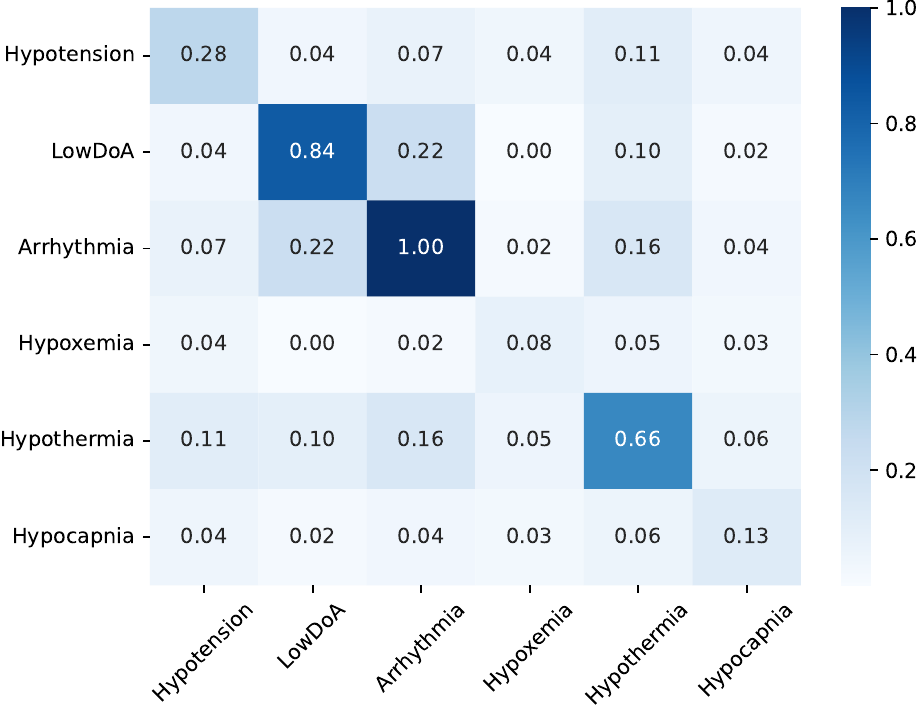}
    \caption{Co-occurrence matrix of adverse events}
    \label{fig: co matric}
\end{figure}

To analyze the optimization properties of LCRLoss, we further conducted the loss gradients. The gradient curves of different loss functions are depicted in~\textbf{Fig.~\ref{fig: Gradients}}. While conventional losses (e.g., BCE, FL) exhibit vanishing gradients for high-confidence predictions, LCRLoss maintains informative gradients, enabling effective optimization even in confident regions. Notably, LCRLoss promotes similar logits for frequently co-occurring events. For example, arrhythmia shares consistent gradient trends with low DoA and hypothermia. In contrast to, hypoxemia and hypocapnia display more independent patterns. This aligns with the label co-occurrence matrix shown in~\textbf{Fig.~\ref{fig: co matric}}, highlighting LCRLoss’s capacity to capture inter-label dependencies and enhance recall in multi-label classification.

\section{Conclusion}
In this work, we propose a novel multi-label classification framework for early warning of adverse events based on the proposed MuAE dataset. By integrating static covariates and dynamic vital signs via the TAFiLM module and leveraging the Transformer architecture to capture complex temporal dependencies, IAENet effectively models heterogeneous clinical time series. To address label imbalance and inter-label dependencies, we introduce a label-constrained reweighting loss that combines batch-wise dynamic weighting and label co-occurrence constraints.

Experimental results show that our approach consistently outperforms strong baselines across multiple evaluation metrics, highlighting its potential for clinical risk prediction. In future work, we plan to incorporate external knowledge and evaluate the framework on broader clinical datasets and event types. This study offers a foundation for developing intelligent early warning systems to support real-time decision-making in perioperative care.

\section*{Acknowledgments}
This work was supported in part by the Sichuan Provincial Science and Technology Department (No.2022YFQ0108) and in part by the Chengdu Science and Technology Program (No.2022-YF04-00078-JH).


\bibliography{aaai2026}

@article{nepogodiev2019global,
  title={Global burden of postoperative death},
  author={Nepogodiev, Dmitri and Martin, Janet and Biccard, Bruce and Makupe, Alex and Bhangu, Aneel and Nepogodiev, D and others},
  journal={Lancet},
  volume={393},
  number={10170},
  pages={401},
  year={2019}
}

@article{varghese2024artificial,
  title={Artificial intelligence in surgery},
  author={Varghese, Chris and Harrison, Ewen M and O’Grady, Greg and Topol, Eric J},
  journal={Nature medicine},
  volume={30},
  number={5},
  pages={1257--1268},
  year={2024},
  publisher={Nature Publishing Group US New York}
}

@article{meara2015global,
  title={Global Surgery 2030: evidence and solutions for achieving health, welfare, and economic development},
  author={Meara, John G and Leather, Andrew JM and Hagander, Lars and Alkire, Blake C and Alonso, Nivaldo and Ameh, Emmanuel A and Bickler, Stephen W and Conteh, Lesong and Dare, Anna J and Davies, Justine and others},
  journal={The lancet},
  volume={386},
  number={9993},
  pages={569--624},
  year={2015},
  publisher={Elsevier}
}

@article{cai2025advances,
  title={Advances in automated anesthesia: a comprehensive review},
  author={Cai, Xiuding and Wang, Xueyao and Zhu, Yaoyao and Yao, Yu and Chen, Jiao},
  journal={Anesthesiology and Perioperative Science},
  volume={3},
  number={1},
  pages={3},
  year={2025},
  publisher={Springer}
}

@article{wang2024deep,
  title={Deep time series models: A comprehensive survey and benchmark},
  author={Wang, Yuxuan and Wu, Haixu and Dong, Jiaxiang and Liu, Yong and Long, Mingsheng and Wang, Jianmin},
  journal={arXiv preprint arXiv:2407.13278},
  year={2024}
}

@article{cai2024outcomes,
  title={Outcomes of clinical decision support systems in real-world perioperative care: a systematic review and meta-analysis},
  author={Cai, Jianwen and Li, Peiyi and Li, Weimin and Zhu, Tao},
  journal={International Journal of Surgery},
  volume={110},
  number={12},
  pages={8057--8072},
  year={2024},
  publisher={LWW}
}

@article{hwang2023intraoperative,
  title={Intraoperative hypotension prediction based on features automatically generated within an interpretable deep learning model},
  author={Hwang, Eugene and Park, Yong-Seok and Kim, Jin-Young and Park, Sung-Hyuk and Kim, Junetae and Kim, Sung-Hoon},
  journal={IEEE Transactions on Neural Networks and Learning Systems},
  volume={35},
  number={10},
  pages={13887--13901},
  year={2023},
  publisher={IEEE}
}

@article{lee2021deep,
  title={Deep learning models for the prediction of intraoperative hypotension},
  author={Lee, Solam and Lee, Hyung-Chul and Chu, Yu Seong and Song, Seung Woo and Ahn, Gyo Jin and Lee, Hunju and Yang, Sejung and Koh, Sang Baek},
  journal={British journal of anaesthesia},
  volume={126},
  number={4},
  pages={808--817},
  year={2021},
  publisher={Elsevier}
}

@article{moon2024frequency,
  title={Frequency domain deep learning with non-invasive features for intraoperative hypotension prediction},
  author={Moon, Jeong-Hyeon and Lee, Garam and Lee, Seung Mi and Ryu, Jiho and Kim, Dokyoon and Sohn, Kyung-Ah},
  journal={IEEE Journal of Biomedical and Health Informatics},
  volume={28},
  number={10},
  pages={5718--5728},
  year={2024},
  publisher={IEEE}
}

@inproceedings{yang2024dynamic,
  title={Dynamic Prediction of Intraoperative Hypotension Based on Hemodynamic Monitoring Data With a Transformer-Based Deep Learning Model},
  author={Yang, Kai and Ren, Mucheng and Xu, Jun and Zeng, Xian},
  booktitle={2024 IEEE International Conference on Bioinformatics and Biomedicine (BIBM)},
  pages={5166--5173},
  year={2024},
  organization={IEEE}
}

@inproceedings{lu2023composite,
  title={A composite multi-attention framework for intraoperative hypotension early warning},
  author={Lu, Feng and Li, Wei and Zhou, Zhiqiang and Song, Cheng and Sun, Yifei and Zhang, Yuwei and Ren, Yufei and Liao, Xiaofei and Jin, Hai and Luo, Ailin and others},
  booktitle={Proceedings of the AAAI Conference on Artificial Intelligence},
  volume={37},
  pages={14374--14381},
  year={2023}
}

@article{chen2021forecasting,
  title={Forecasting adverse surgical events using self-supervised transfer learning for physiological signals},
  author={Chen, Hugh and Lundberg, Scott M and Erion, Gabriel and Kim, Jerry H and Lee, Su-In},
  journal={NPJ digital medicine},
  volume={4},
  number={1},
  pages={167},
  year={2021},
  publisher={Nature Publishing Group UK London}
}

@article{park2023machine,
  title={Machine learning-based prediction of intraoperative hypoxemia for pediatric patients},
  author={Park, Jung-Bin and Lee, Ho-Jong and Yang, Hyun-Lim and Kim, Eun-Hee and Lee, Hyung-Chul and Jung, Chul-Woo and Kim, Hee-Soo},
  journal={PLoS One},
  volume={18},
  number={3},
  pages={e0282303},
  year={2023},
  publisher={Public Library of Science San Francisco, CA USA}
}

@article{lundberg2018explainable,
  title={Explainable machine-learning predictions for the prevention of hypoxaemia during surgery},
  author={Lundberg, Scott M and Nair, Bala and Vavilala, Monica S and Horibe, Mayumi and Eisses, Michael J and Adams, Trevor and Liston, David E and Low, Daniel King-Wai and Newman, Shu-Fang and Kim, Jerry and others},
  journal={Nature biomedical engineering},
  volume={2},
  number={10},
  pages={749--760},
  year={2018},
  publisher={Nature Publishing Group UK London}
}

@article{lee2019data,
  title={Data driven investigation of bispectral index algorithm},
  author={Lee, Hyung-Chul and Ryu, Ho-Geol and Park, Yoonsang and Yoon, Soo Bin and Yang, Seong Mi and Oh, Hye-Won and Jung, Chul-Woo},
  journal={Scientific reports},
  volume={9},
  number={1},
  pages={13769},
  year={2019},
  publisher={Nature Publishing Group UK London}
}

@article{lee2018prediction,
  title={Prediction of bispectral index during target-controlled infusion of propofol and remifentanil},
  author={Lee, Hyung-Chul and Ryu, Ho-Geol and Chung, Eun-Jin and Jung, Chul-Woo},
  journal={Anesthesiology},
  volume={128},
  number={3},
  pages={492--501},
  year={2018},
  publisher={Wolters Kluwer}
}

@inproceedings{yeche2023temporal,
  title={Temporal label smoothing for early event prediction},
  author={Y{\`e}che, Hugo and Pace, Aliz{\'e}e and Ratsch, Gunnar and Kuznetsova, Rita},
  booktitle={International Conference on Machine Learning},
  pages={39913--39938},
  year={2023},
  organization={PMLR}
}

@inproceedings{liu2024itransformer,
title={iTransformer: Inverted Transformers Are Effective for Time Series Forecasting},
author={Yong Liu and Tengge Hu and Haoran Zhang and Haixu Wu and Shiyu Wang and Lintao Ma and Mingsheng Long},
booktitle={The Twelfth International Conference on Learning Representations},
year={2024},
url={https://openreview.net/forum?id=JePfAI8fah}
}

@inproceedings{zhou2021informer,
  title={Informer: Beyond efficient transformer for long sequence time-series forecasting},
  author={Zhou, Haoyi and Zhang, Shanghang and Peng, Jieqi and Zhang, Shuai and Li, Jianxin and Xiong, Hui and Zhang, Wancai},
  booktitle={Proceedings of the AAAI conference on artificial intelligence},
  volume={35},
  pages={11106--11115},
  year={2021}
}

@inproceedings{wu2023timesnet,
title={TimesNet: Temporal 2D-Variation Modeling for General Time Series Analysis},
author={Haixu Wu and Tengge Hu and Yong Liu and Hang Zhou and Jianmin Wang and Mingsheng Long},
booktitle={The Eleventh International Conference on Learning Representations },
year={2023},
url={https://openreview.net/forum?id=ju_Uqw384Oq}
}

@inproceedings{zhang2023crossformer,
  title={Crossformer: Transformer utilizing cross-dimension dependency for multivariate time series forecasting},
  author={Zhang, Yunhao and Yan, Junchi},
  booktitle={The eleventh international conference on learning representations},
  year={2023}
}

@inproceedings{zhou2022fedformer,
  title={Fedformer: Frequency enhanced decomposed transformer for long-term series forecasting},
  author={Zhou, Tian and Ma, Ziqing and Wen, Qingsong and Wang, Xue and Sun, Liang and Jin, Rong},
  booktitle={International conference on machine learning},
  pages={27268--27286},
  year={2022},
  organization={PMLR}
}

@inproceedings{nie2022time,
  title = {A Time Series is Worth 64 Words: Long-term Forecasting with Transformers},
  author = {Nie, Yuqi and
               H. Nguyen, Nam and
               Sinthong, Phanwadee and 
               Kalagnanam, Jayant},
  booktitle = {International Conference on Learning Representations},
  year      = {2023}
}

@article{wu2021autoformer,
  title={Autoformer: Decomposition transformers with auto-correlation for long-term series forecasting},
  author={Wu, Haixu and Xu, Jiehui and Wang, Jianmin and Long, Mingsheng},
  journal={Advances in neural information processing systems},
  volume={34},
  pages={22419--22430},
  year={2021}
}

@article{liu2022non,
  title={Non-stationary transformers: Exploring the stationarity in time series forecasting},
  author={Liu, Yong and Wu, Haixu and Wang, Jianmin and Long, Mingsheng},
  journal={Advances in neural information processing systems},
  volume={35},
  pages={9881--9893},
  year={2022}
}

@article{lin2023segrnn,
  title={Segrnn: Segment recurrent neural network for long-term time series forecasting},
  author={Lin, Shengsheng and Lin, Weiwei and Wu, Wentai and Zhao, Feiyu and Mo, Ruichao and Zhang, Haotong},
  journal={arXiv preprint arXiv:2308.11200},
  year={2023}
}

@inproceedings{zeng2023transformers,
  title={Are transformers effective for time series forecasting?},
  author={Zeng, Ailing and Chen, Muxi and Zhang, Lei and Xu, Qiang},
  booktitle={Proceedings of the AAAI conference on artificial intelligence},
  volume={37},
  pages={11121--11128},
  year={2023}
}

@inproceedings{ridnik2021asymmetric,
  title={Asymmetric loss for multi-label classification},
  author={Ridnik, Tal and Ben-Baruch, Emanuel and Zamir, Nadav and Noy, Asaf and Friedman, Itamar and Protter, Matan and Zelnik-Manor, Lihi},
  booktitle={Proceedings of the IEEE/CVF international conference on computer vision},
  pages={82--91},
  year={2021}
}

@inproceedings{lin2017focal,
  title={Focal loss for dense object detection},
  author={Lin, Tsung-Yi and Goyal, Priya and Girshick, Ross and He, Kaiming and Doll{\'a}r, Piotr},
  booktitle={Proceedings of the IEEE international conference on computer vision},
  pages={2980--2988},
  year={2017}
}

@article{chawla2002smote,
  title={SMOTE: synthetic minority over-sampling technique},
  author={Chawla, Nitesh V and Bowyer, Kevin W and Hall, Lawrence O and Kegelmeyer, W Philip},
  journal={Journal of artificial intelligence research},
  volume={16},
  pages={321--357},
  year={2002}
}

@article{lee2022vitaldb,
  title={VitalDB, a high-fidelity multi-parameter vital signs database in surgical patients},
  author={Lee, Hyung-Chul and Park, Yoonsang and Yoon, Soo Bin and Yang, Seong Mi and Park, Dongnyeok and Jung, Chul-Woo},
  journal={Scientific Data},
  volume={9},
  number={1},
  pages={279},
  year={2022},
  publisher={Nature Publishing Group UK London}
}

@inproceedings{perez2018film,
  title={Film: Visual reasoning with a general conditioning layer},
  author={Perez, Ethan and Strub, Florian and De Vries, Harm and Dumoulin, Vincent and Courville, Aaron},
  booktitle={Proceedings of the AAAI conference on artificial intelligence},
  volume={32},
  year={2018}
}

@article{cheung2015predictors,
  title={Predictors of intraoperative hypotension and bradycardia},
  author={Cheung, Christopher C and Martyn, Alan and Campbell, Norman and Frost, Shaun and Gilbert, Kenneth and Michota, Franklin and Seal, Douglas and Ghali, William and Khan, Nadia A},
  journal={The American journal of medicine},
  volume={128},
  number={5},
  pages={532--538},
  year={2015},
  publisher={Elsevier}
}

@article{schnider2021relationship,
  title={Relationship between propofol target concentrations, bispectral index, and patient covariates during anesthesia},
  author={Schnider, Thomas W and Minto, Charles F and Egan, Talmage D and Filipovic, Miodrag},
  journal={Anesthesia \& Analgesia},
  volume={132},
  number={3},
  pages={735--742},
  year={2021},
  publisher={LWW}
}

@article{bahador2021multimodal,
  title={Multimodal spatio-temporal-spectral fusion for deep learning applications in physiological time series processing: A case study in monitoring the depth of anesthesia},
  author={Bahador, Nooshin and Jokelainen, Jarno and Mustola, Seppo and Kortelainen, Jukka},
  journal={Information Fusion},
  volume={73},
  pages={125--143},
  year={2021},
  publisher={Elsevier}
}

@article{cao2019learning,
  title={Learning imbalanced datasets with label-distribution-aware margin loss},
  author={Cao, Kaidi and Wei, Colin and Gaidon, Adrien and Arechiga, Nikos and Ma, Tengyu},
  journal={Advances in neural information processing systems},
  volume={32},
  year={2019}
}

@inproceedings{cui2019class,
  title={Class-balanced loss based on effective number of samples},
  author={Cui, Yin and Jia, Menglin and Lin, Tsung-Yi and Song, Yang and Belongie, Serge},
  booktitle={Proceedings of the IEEE/CVF conference on computer vision and pattern recognition},
  pages={9268--9277},
  year={2019}
}

\end{document}


\appendix
\section{Supplementary A: Dataset Description}
In this section, we provide additional details about the MuAE dataset used in our study. The dataset consists of 873 patients collected from the VitalDB dataset, spanning the perioperative period. 

\subsection{Inclusion and Exclusion Criteria}

To construct a reliable and clinically meaningful dataset, we applied the following inclusion and exclusion criteria:
\begin{itemize}
    \item Surgery duration longer than 2 hours
    \item General anesthesia (N = 4,630)
    \item Age $\geq$ 18 years old
    \item Weight $>$ 35 kg
    \item ASA $<$ 6
\end{itemize}

After applying these criteria, 1,311 surgeries were included for the feature selection. The statistical information regarding the dataset division is presented in~\textbf{Table~\ref {tab: dataset information}}.

\subsection{Feature Selection}
As shown in~\textbf{Table~\ref{tab: dataset variables}}, we then selected 20 time-series features relevant to intraoperative monitoring. These include:
\begin{itemize}
    \item Demographic Data: 'AGE', 'SEX', 'WEIGHT', 'HEIGHT', 'ASA'
    \item Infusion pump volume and concentration parameters(e.g., 'Orchestra/PPF20\_VOL', 'Orchestra/PPF20\_CE', 'Orchestra/RFTN20\_VOL', 'Orchestra/RFTN20\_CE')
    \item BIS monitoring ('BIS/BIS')
    \item Other relevant channels related to anesthetic administration and physiological monitoring
\end{itemize}

Among these, the demographic data serve as constant features and act as covariates in predicting other variables.

\begin{figure}[h]
    \centering
    \includegraphics[width=0.95\linewidth]{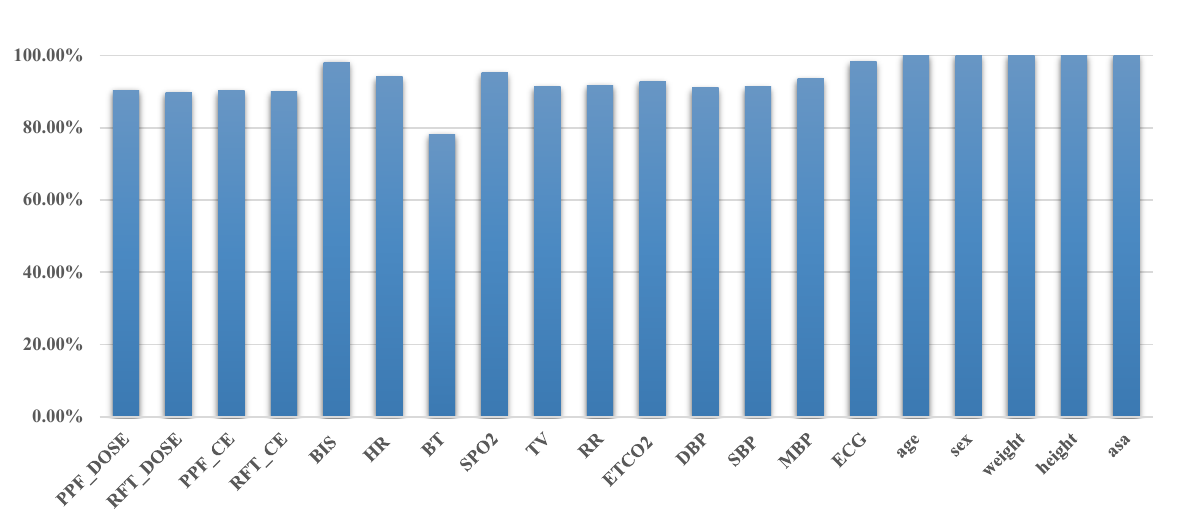}
    \caption{Non-Missing rate for variables in the MuAE dataset}
    \label{fig: missing-label}
\end{figure}

Finally, the data were then preprocessed via resampling at 2-second intervals, handling missing values, converting infusion volumes to rates, and filtering based on BIS signal thresholds. As shown in~\textbf{Fig.~\ref{fig: missing-label}}, the MuAE dataset exhibits varying degrees of missingness across different variables, with missing rates ranging from 0\% to 25\%. After preprocessing, 873 cases were retained and split into training (70\%), test (10\%), and validation (20\%) sets for model development and evaluation. This processed MuAE dataset serves as the foundation for model training, evaluation, and ablation studies.


\begin{table*}
    \centering
    \begin{tabular}{lccccc}
    \toprule
     & Training dataset & Validation dataset & Test dataset & \multirow{2}{*}{P-value} & \multirow{2}{*}{Test}\\
     & (n = 610, 70\%)  & (n = 88, 10\%) & (n = 175, 20\%) & \\
    \midrule
    Age, years $\pm$SD   & 59.59 $\pm$ 14.15  & 61.22 $\pm$ 15.48 &  60.88 $\pm$ 12.98 & 0.502 & Kruskal-Wallis\\
    Sex, n(\%) &&&&0.776 & Chi-squared     \\
    Female  & 257 (42.1) & 38 (43.2) & 79 (45.1) & \\
    Male & 353 (57.9) & 50 (56.8) & 96 (54.9) &\\
    Weight    & 62.94 $\pm$ 11.99 & 62.87 $\pm$ 11.76 & 61.53 $\pm$ 10.57 & 0.379 & Kruskal-Wallis \\
    Height    & 163.75 $\pm$ 8.25 & 163.38 $\pm$ 9.49 & 163.07 $\pm$ 8.46 & 0.586 & Kruskal-Wallis\\
    ASA, n(\%) & & & & 0.629  & Chi-squared  \\
    I    & 123 (20.2) &13 (14.8)   &42 (24.0) &\\
    II   & 418 (68.5) &64 (72.7)  &117 (66.9)&\\
    III  & 67 (11.0)  &11 (12.5)  &16 (9.1)&\\
    IV   & 2 (0.3)   &0 (0.0)   &0 (0.0)&\\
    V    & 0 (0.0)  &0 (0.0)   &0 (0.0)&\\
      
    \bottomrule
    \end{tabular}
    \caption{Baseline characteristics of the dataset}
    \label{tab: dataset information}
\end{table*}

\begin{table*}
    \centering
    \begin{tabular}{lcc}
    \toprule
    Device/Variable & Description & Covariate\\
    
    \midrule
    Age & Age of the patient & $\checkmark$ \\
    Sex & Gender of the patient & $\checkmark$ \\
    Weight & Body weight of the patient & $\checkmark$ \\
    Height & Height of the patient & $\checkmark$\\ 
    ASA & ASA classification & $\checkmark$ \\
    Orchestra/PPF20\_VOL & Propofol infusion volume \\
    Orchestra/RFTN20\_VOL& Remifentanil infusion volume \\
    Orchestra/PPF20\_CE& Propofol effect-site concentration \\
    Orchestra/RFTN20\_CE& Remifentanil effect-site concentration \\
    Solar8000/HR& Heart rate \\
    Solar8000/BT& Body temperature \\
    Solar8000/ART\_DBP& Arterial diastolic blood pressure \\
    Solar8000/ART\_SBP& Arterial systolic blood pressure \\
    Solar8000/ART\_MBP& Arterial mean blood pressure \\
    Solar8000/ETCO2& End-tidal carbon dioxide \\
    Solar8000/PLETH\_SPO2& Peripheral oxygen saturation \\
    Solar8000/VENT\_TV & Measured tidal volume (from ventilator) \\
    Solar8000/VENT\_RR & Respiratory rate (from ventilator) \\
    BIS/BIS& Bispectral index \\
    SNUADC/ECG\_II &  ECG lead II wave\\
     
    \bottomrule
    \end{tabular}
    \caption{Summary of Variables Selected from the MuAE Dataset}
    \label{tab: dataset variables}
\end{table*}

\section{Supplementary B: LCRLoss}
\begin{figure}[h]
    \centering
    \includegraphics[width=1\linewidth]{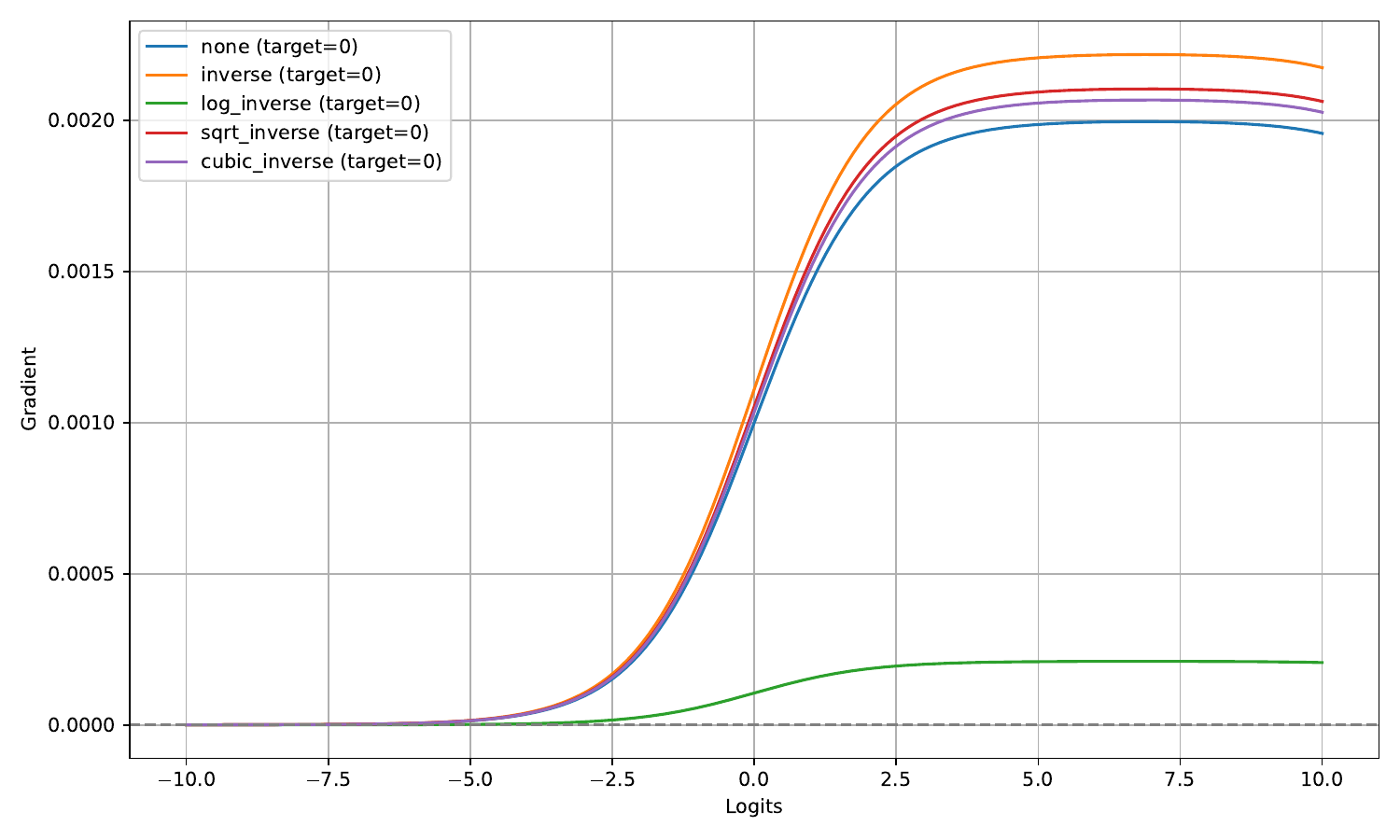}
    \caption{Gradient Analysis for Negative Sample}
    \label{fig: loss Gradient Analysis -}
\end{figure}

\begin{figure}[h]
    \centering
    \includegraphics[width=1\linewidth]{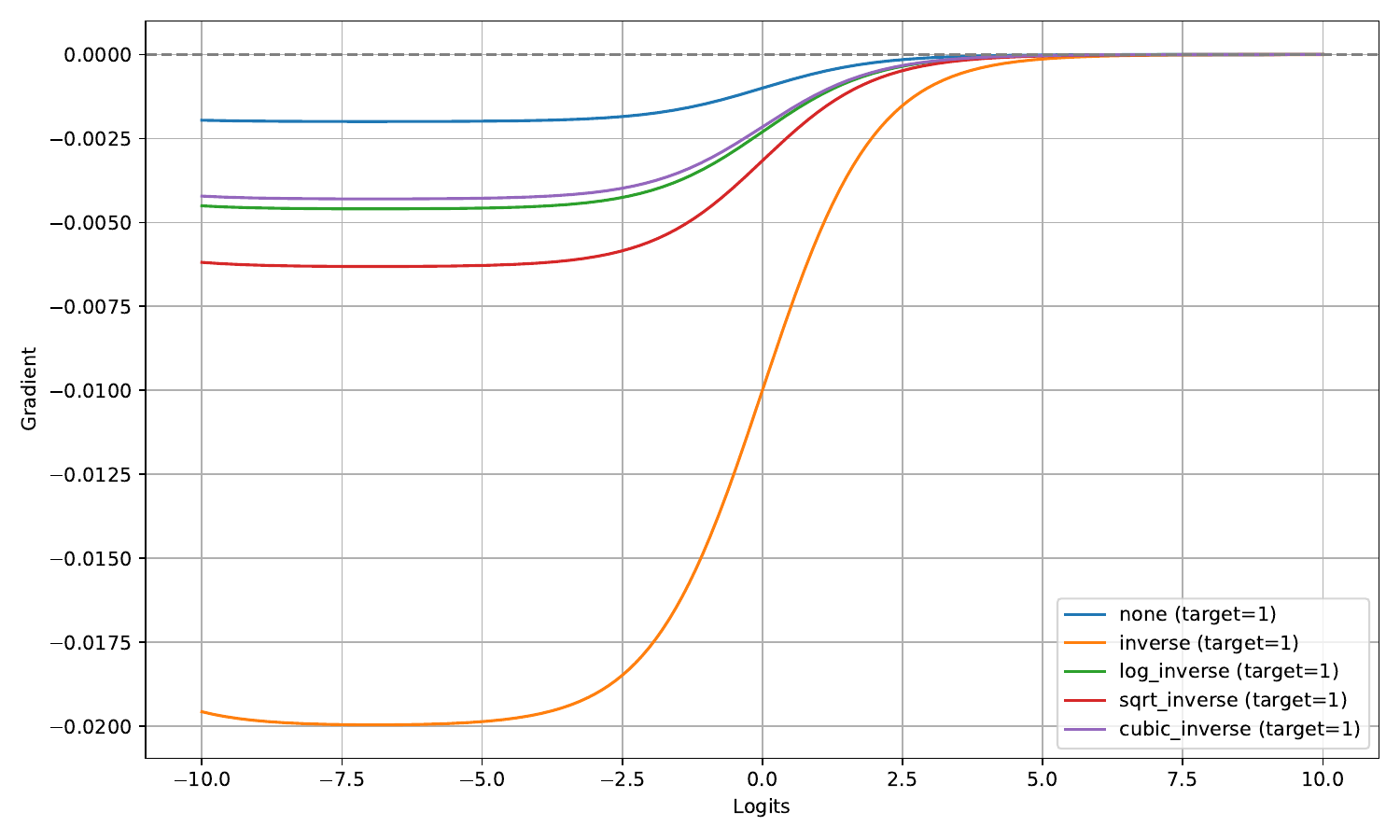}
    \caption{Gradient Analysis for Positive Sample}
    \label{fig: loss Gradient Analysis +}
\end{figure}

We conducted supplementary experiments to further evaluate the effectiveness of the proposed label frequency reweighting strategies. Overly large weights may cause the model to over-optimize for recall, leading to more false positives and a decrease in precision. To balance this trade-off, we visualize the gradient curves under different weighting strategies, as shown in~\textbf{Fig.~\ref{fig: loss Gradient Analysis -}} and~\textbf{Fig.~\ref{fig: loss Gradient Analysis +}}, which depict how gradients change for positive samples (target=1) and negative samples (target=0). The results show that the inverse weighting strategy produces the largest gradients and the fastest learning, but it also tends to cause overfitting. In contrast, the $sqrt\_inverse$ strategy offers a more balanced solution, alleviating class imbalance while effectively controlling the risk of gradient explosion.

